\documentclass{article}
\usepackage{arxiv}

\usepackage[utf8]{inputenc} 
\usepackage[T1]{fontenc}    
\usepackage{hyperref}       
\usepackage{url}            
\usepackage{booktabs}  
\usepackage[utf8]{inputenc}
\usepackage{array}
\usepackage{wrapfig}
\usepackage{multirow}
\usepackage{tabu}
\usepackage{fancyhdr}
\usepackage[british]{babel}
\usepackage{hhline}
\usepackage{multirow}
\usepackage{amsfonts}       
\usepackage{nicefrac}       
\usepackage{microtype}      
\usepackage{lipsum}
\usepackage{bbm}
\usepackage{amsmath}
\usepackage{graphicx}
\graphicspath{{./}}
\title{Exploring and Improving Robustness of Multi Task Deep Neural Networks via Domain Agnostic Defenses}

\author{
   Kashyap Coimbatore Murali \\
  High School Student\\
  West Windsor-Plainsboro High School South\\
  West Windsor, NJ 08550 \\
  \texttt{katchu11@gmail.com} \\
}

\begin{document}
\maketitle

\begin{abstract}
In this paper, we explore the robustness of the Multi-Task Deep Neural Networks (MT-DNN) against non-targeted adversarial attacks across Natural Language Understanding (NLU) tasks as well as some possible ways to defend against them. Liu et al., have shown that the Multi-Task Deep Neural Network \cite{liu2019multitask}, due to the regularization effect produced when training as a result of it's cross task data, is more robust than a vanilla BERT model trained only on one task (1.1\%-1.5\% absolute difference). We further show that although the MT-DNN has generalized better, making it easily transferable across domains and tasks, it can still be compromised as after only 2 attacks (1-character and 2-character) the accuracy drops by 42.05\% and 32.24\% for the SNLI and SciTail tasks. Finally, we propose a domain agnostic defense which restores the model's accuracy (36.75\% and 25.94\% respectively) as opposed to a general-purpose defense or an off-the-shelf spell checker.
\end{abstract}

\keywords{Model Robustness \and Domain Adaptation \and Adversarial Defense}

\section{Introduction}


The pace of deep learning research in NLU tasks is extremely fast, with new papers being published consistently pushing the benchmark for various tasks in the field. One such example is the Multi-Task Deep Neural Network (MT-DNN) \cite{liu2019multitask}, a hybrid model which utilizes a combination of language model pretraining along with multi task learning in order to improve the learning of semantic representations to augment the performance across various NLU tasks.

Even with the rapid pace by which benchmarks are pushed across supervised learning tasks using deep learning, many of these models are susceptible to attacks which are almost imperceptible to humans. After it was proven that image classifying neural networks can be fooled to predict incorrect classes by adding some noise that was unidentifiable by humans \cite{szegedy2013intriguing}, the focus began to shift on how NLU models can be attacked to make false predictions on sentences which are obviously different classes when observed by humans.

There are four fundamental ways of creating adversarial text examples: adding, deleting, swapping adjacent characters, or replacing characters with those that are closer on the keyboard \cite{Li_2019}. We explore various combinations of these character level \emph{swap} attacks on the perturbed sentences so that the text is falsely classified when fed through the MT-DNN, yet easily comprehensible by humans. These alterations are founded through psycho-linguistic studies, which show that the semantic meaning of the words can be easily comprehended even if the characters are jumbled as long as the first and last characters remain the same (Ex: \emph{Cmabridge Uinversity}).

We proceed to explore various methods by which defenses can be introduced such that even if perturbations exist within multiple characters across a sentence, the accuracy can be restored. We utilize a robust word recognition model inspired from Semicharacter RNNs \cite{sakaguchi2016robsut} which can correct a word by generating encodings for the first character, then treating the middle characters as a bag of characters where frequency is the feature as opposed to position, and finally generating a final encoding for the last character. We further propose a \emph{multi-task defense} where the word recognition model is pretrained off a combination of multiple tasks (similar to how the MT-DNN is trained), which we can then fine-tune  to the task at hand, thus ensuring a modular defense which can learn the intricacies of the domain at which it's operating in.

\section{Related Works}
\subsection{Pre-trained Language Models}
There are various pre-trained language models which employ different strategies to apply them to downstream tasks such as feature-based method (ELMo \cite{Peters:2018}) or a fine-tuning method (XLNet \cite{yang2019xlnet}, BERT \cite{BERT}, Open AI GPT \cite{radford2019language} in order to achieve state-of-the-art accuracy in different NLI tasks. These models employ the large swaths of unlabelled textual data in order to train contextualized word representations, thus eliminating the need to create bulky task specific architectures.
\subsection{Application of Pretrained Models on Downstream Tasks}
Due to the extent of the pretraining and the complexity within these models, in order to make a prediction we simply need to add a classification layer at the end to make a prediction on different tasks (as mentioned later). However in order to achieve higher accuracies  need to incorporate the data of a given domain. There are three main ways to accomplish this: finetuning the entire mode along with the classification layer, fine tuning  the BERT \cite{BERT} model using Masked Language Modeling, and Next Sentence Prediction (essentially not using the classification layer), and Multi Task Finetuning. In the original approach we simply add a classification layer based on the task and fine tune \emph{all} the model's parameters using the in domain data for that use case in the second approach we don't add the classification layer but simply retrain the BERT \cite{BERT} model using it's original loss functions for the Masked Language Modeling, and next Sentence Prediction Tasks using the in domain data, thus allowing it to create domain-specific contextual embeddings prior o the application of the classification layer. The final approach ascertains the combination of multiple tasks by conjoining the entire dataset across tasks to create a generalized model, while using task specific loss functions when applying it to a specific use-case.
\subsection{Adversarial Attacks}
Adversarial examples can be generated using black-box or white-box
techniques based off of the knowledge of the neural network. Black-box attacks are used when the general architectures, parameters, and hyper-parameters of the model aren't accessible (such as the attack on some foreign API service). White box attacks require complete knowledge of the parameters and architecture of the model to be generated to be generated. 
\newline\newline
These type of adversarial attacks with respect to Natural Language, character or word-level, have been introduced in a variety of manners which primarily either focus on the existence of predominant weaknesses \cite{existence}, through white-box defensive strategies \cite{whitebox}, through black-box strategies \cite{blackbox}, or a combination of both \cite{Li_2019} \cite{survey}
\newline\newline
There also exist different types of perturbations that can be performed on the text including: adding, swapping, deleting, or inserting characters which are positionally closer on the keyboard. These attacks are shown in \cite{cost} they have the ability to significantly decrease words per minute (wpm) by up to 36\% based on the location of the attack and that although 50\% of survey respondents didn't understand a few words, they were still able to answer the perturbed questions with a high accuracy. 
\subsection{Defenses}
In order to defend against gradient based attacks (white-box) or non-gradient based attacks (black-box) alike, one potential method is through the use of spelling correction. Spelling correction \cite{Kukich1992TechniquesFA} is often viewed as a sub-task of grammatical error correction \cite{ng-etal-2013-conll} \cite{DBLP:journals/corr/SchmaltzKRS17}. Classical methods utilize a body of text which includes the number of times the word is used, then analyze different errors and finally compute the Levenshtein distance between the chosen word and the target word. Another approach was through the use of a pretrained language embedding trained to identify n-grams in order to make a prediction based off of the previous word. Lately, deep learning inspired approaches have been introduced, which analyze both the context and orthography of the input. There are also off-the-shelf spell checkers such as AHD, or Hunspell check which are trained off of a general purpose dataset and allow for ease of use due to their packaging as Python libraries. This work utilizes the SemiCharacter RNNs \cite{sakaguchi2016robsut}
\label{sec:headings}





\section{Approach}
\subsection{MT-DNN Model Architecture}
\begin{figure}
    \centering
    \includegraphics[width = \textwidth]{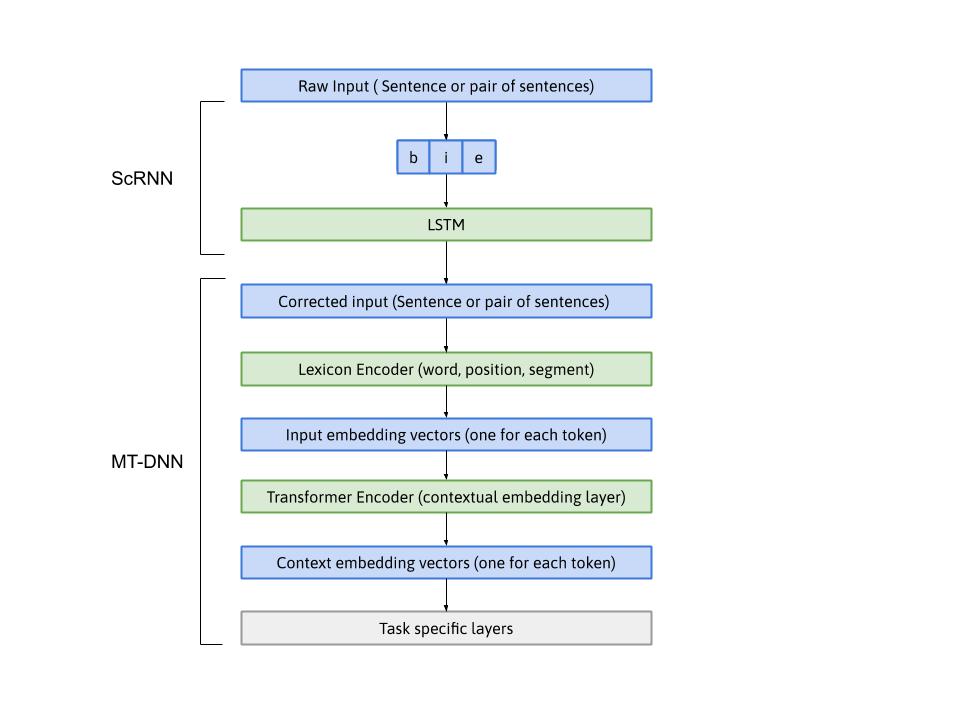}
    \caption{This is the architecture of the model including the ScRNN and the MT-DNN. This shows how an input flows through the ScRNN, gets corrected with the LSTM, and then flows through the model from the lexicon encoder to the transformer encoder to generate the contextual word embedding after which a task specific output layer exists to make a prediction. Image combined elements from \cite{BERT} and \cite{sakaguchi2016robsut} }
    \label{fig:mtdnn}
\end{figure}
The model architecture contains a chain of shared text encoding layers which has  of a lexicon encoder and a contextual transformer encoder, which are trained and shared across all tasks.
The shared lexicon and contextual transformer encoder is initialized using the BERT \cite{BERT} model as released by huggingface's pytorch library \cite{wolf2019huggingfaces}. These parameters are then update during the Multi Task Learning (MTL) phase of creating a pretrained MT-DNN model \cite{liu2019multitask}. The output layers are then appended to the bottom as shown in (Figure 1), which are chosen based off the task. The individual components are further elaborated below.

\subsubsection{Lexicon Encoder} 
The lexicon encoder is the pre-processing step in order to generate the contextual embedding from the Transformer Encoder [16]. Given an input X, it is first tokenized using WordPiece tokenization \cite{wu2016googles} which splits a token into pieces that model can understand (Ex: "I am playing" -> ["i", "am", "play","\#\#ing"]), with a word that it may not know at all being split all the way down to the character level, hence ensuring that there would never be an OOV error for words created from characters of the alphabet of the language for which the model is trained. In exceptional cases where the character isn't recognized, the token is replaced with a [UNK] token, thus ensuring that no error is produced. The first token of the sequence is always the classification token ([CLS]) with the final hidden state of this token containing the representation of the entire sequence and is used for sentence level classification tasks. In situations where there are a pair of sentences (2 sentence classification tasks), the sentences are separated with the [SEP] token which denotes what segment it belongs to. This would result in a list of tokens like this:
\newline
\newline
One sentence classification:
\newline
\newline
["[CLS]","i","am","play","\#\#ing"] 
\newline
\newline
Two sentence classification task:  
\newline
\newline
["[CLS]","i","am",play,"\#\#ing","[SEP]","i","am","talk","\#\#ing","[SEP]"] 
\newline
\newline
This is then mapped to a sequence of input embedding vectors for each token which is created by summing the corresponding word, segment, and positional embeddings (Figure 2).
\begin{figure}
    \centering
    \includegraphics[width = 0.75\textwidth]{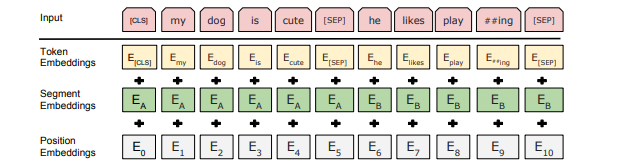}
    \caption{This shows how the lexicon encoder creates an input representation by summing the token, segment, and position embeddings from a given input. From \cite{BERT}}
    \label{fig:bert}
\end{figure}

\subsubsection{Transformer Encoder} 
The multi-layer bidirectional transformer-encoder \cite{vaswani2017attention} layer then maps the input representation vectors to a sequence of contextual embedding vectors. This becomes the representation which is used across all tasks. The transformer layer utilizes
a self-attention mechanism which directly models contextual relationships across all tokens in a sequence. A key difference in the MT-DNN model as opposed to the vanilla BERT \cite{BERT} model is the utilization of Multi Task Objectives (Figure 3) as well as pre-training.
\newline

\subsubsection{Single-Sentence Classification Task}

Given the contextual embedding [CLS] token(the aggregate semantic representation of the entire sentence), it can be used to make a prediction regarding the sentence as a whole. The output probability is based on a probability distribution of size $n_{labels}$. 
The probability is predicted using a softmax function over the linear output layer.
\begin{equation}
    P(c|x) = \textup{softmax}(W_\emph{task}*x)
\end{equation}
The equation represents the probability of each class given a contextual embedding input x, where W\_task represents the learned weight matrix for a given task.

The loss function for this task is binary cross entropy loss as the objective.
\begin{equation}
  -\sum_c\mathbbm{1}(X,c)\textup{log}(P_r(c|X))
\end{equation}
where $\mathbbm{1}(X,c)$ is either 0 or 1 (binary classification) if
the predicted class label $c$ is the correct classification for $X$(single sentence), penalized by a factor of $P_r$ (softmax equation)

\emph{Input: "The movie was great and had great actors and actresses"
      Output: Positive Review
}
\subsubsection{Pairwise Text Classification}

In the pairwise text classification problem, the objective is to output a label given an input of 2 separate sentences. These sentences are packed together into one input sequence which includes a premise and the hypothesis(2 separate components). The original MT-DNN \cite{liu2019multitask} utilizes a Stochastic Answer Network [18] as opposed to just predict a label which allows it to  maintain a state and iteratively refines its predictions for k-number of steps (where k is a hyperparameter) and averages the prediction at each step k to create a final prediction which improves the robustness of the model.

\begin{equation}
    P_r^k = \textup{softmax}(W_\emph{task}*s^k*x^k)
\end{equation}
The equation above is very similar to the single sentence text prediction however it maintains a state $s$ throughout each step $k$ after which the probability distribution $P_r$ is averaged to produce the final output.

The loss function for this task is binary cross entropy loss.
\begin{equation}
  -\sum_c\mathbbm{1}(X,c)\textup{log}(P_r(c|X))
\end{equation}
where $\mathbbm{1}(X,c)$ is either 0 or 1 (binary classification) if
the predicted class label $c$ is the correct classification for $X$(pair of sentences), penalized by a factor of $P_r$ (softmax equation)

\emph{Input: "A black race car starts up in front of a crowd of people." (Sentence A), "A man is driving down a lonely road." (Sentence B)
\newline
Output: Contradiction
}

\subsubsection{Text Similarity Task}
Given the contextual embedding [CLS] token (the aggregate semantic representation of the entire sequence), it can be used to make a prediction regarding the sequence as a whole. A similarity score can be calculated using:
\begin{equation}
    \textup{Sim}(X_1,X_2) = W_{task}*x
\end{equation}
where $(X_1,X_2)$ represents the 2 different input sequences, $W_{task}$ represents the learned weight matrix for this task, and where $x$ represents the contextual embedding created by the transformer encoder after being fed the input from the lexicon encoder. The function $\textup{Sim}(X_1,X_2)$ can be used to represent the similarity as a real value in the range of $(-\infty,\infty)$

The loss function for this task is mean squared error.
\begin{equation}
    (y-\textup{Sim}(X_1,X_2))^2
\end{equation}
where y represents the ground truth similarity of the text from a range of $(-\infty,\infty)$.

\emph{
Input: "Thank you very much, Commissioner."(Sentence A)	"Thank you very much, Mr Commissioner." (Sentence B)
\newline
Output: 4.500
}
\subsubsection{Relevance Ranking}
Given the contextual embedding [CLS] token (the aggregate semantic representation of the entire sequence of a pair of question and its candidate answer $(Q, A)$. A relevance score can be computed as:
\begin{equation}
    \textup{Rel}(Q,A) = g(W_{task}*x)
\end{equation}
Where a given $Q$ and input representation $x$, we rank all of its candidate answers using the relevancy score.

The loss is similar to the 2 sentence classification task, but instead however reduces the negative log likelihood of the positive answer given questions from the training data.

\begin{equation}
    -\sum_{(Q,A^+)}P_r(A^+|Q)
\end{equation}
where $P_r$ is:
\begin{equation}
    P_r(A^+|Q) = \frac{\textup{exp}(Rel(Q,A^+))}{\sum_{A'\in A}\textup{exp}(Rel(Q,A')}
\end{equation}

\emph{
Input:
\newline
Question: Who edited Electrical World magazine?
\newline
Answers:
\newline
\begin{itemize}
    \item In 1888, the editor of Electrical World magazine was Thomas Commerford Martin
    \item Not all cells in a multicellular plant contain chloroplasts.
\end{itemize}
Output: The first answer is the correct answer for the given query
}

\subsection{Semicharacter RNN (ScRNN) architecture}
The ScRNN \cite{sakaguchi2016robsut} is inspired from findings of the Cmabrigde Uinervtisy effect (also known as typoglycemia), which states that as long as the first and last characters of a word are the same the order of the middle characters doesn't matter in order to know the meaning of the word. This is demonstrated in the paragraph below:
\newline
\newline
\emph{Aoccdrnig to a rscheearch at Cmabrigde Uinervtisy,
it deosn’t mttaer in waht oredr the ltteers in a wrod
are, the olny iprmoetnt tihng is taht the frist and lsat
ltteer be at the rghit pclae. The rset can be a toatl
mses and you can sitll raed it wouthit porbelm. Tihs
is bcuseae the huamn mnid deos not raed ervey lteter
by istlef, but the wrod as a wlohe.
}
\newline
\newline
The ScRNN is shown to significantly increase performance in word spelling correction compared to existing spelling checkers and
charCNNs [19]. For regular RNNs, the input vector is the word or character representations, but for input vector for ScRNNs consist of 3 subvectors
\begin{align}
    x_n &= \begin{bmatrix}
           b_{n} \\
           i_{n} \\
           e_{n}
         \end{bmatrix}
\end{align}
The first and last subvector represents the one hot vector of the first and last characters respectively, and the middle subvector contains the number of occurrences each character has regardless of their order. All three subvectors would be of the size of the alphabet. This concatenation of the subvectors forms $x_n$ which is used as the input to the model which would output the most probably word. The ScRNN performs better as compared to charCNNs from (Kim et al.) \cite{sakaguchi2016robsut} and commercial spell checkers for all the noises that it was tested on (adding - 3.52\% absolute improvement, deleting - 13.89\% absolute improvement, and jumbling - 41.85\% absolute improvement). The difference in performance is especially visualized in jumbling, as oftentimes the other models aren't designed for severely jumbled input. The ScRNN is trained using cross entropy loss as shown below:
\begin{equation}
    y_n = \frac{\textup{exp}(W_h*h_n)}{\sum_{v}\textup{exp}(W_h*h_n)}
\end{equation}
The model learns the weight matrices $W$ in order to predict the word based off the hidden state $h$, with a fixed layer vocabulary of $v$.

\begin{figure}
    \centering
    \includegraphics[width = 0.5\textwidth]{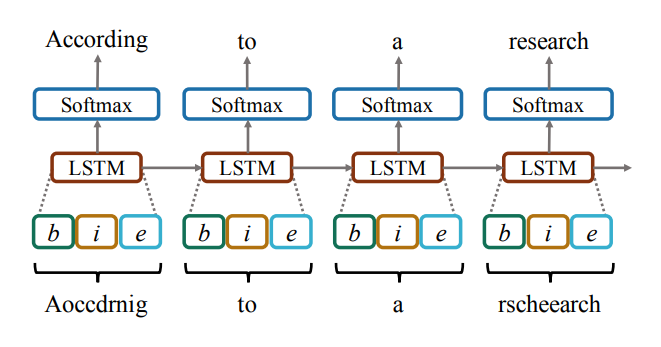}
    \caption{Representation of how the ScRNN takes a misspelled word and uses an LSTM to make a prediction on the correct word. From: \cite{sakaguchi2016robsut}}
    \label{fig:scrnn}
\end{figure}

\clearpage
\subsection{Training Technique}
The training of the MT-DNN \cite{liu2019multitask} is split into two aspects: pretraining off of the GLUE dataset, and finetuning to the task at hand.

\subsubsection{Pretraining}
\textbf{MT-DNN:}
\newline
As opposed to training off one dataset at a time, all the GLUE datasets are merged and shuffled, after which a mini-batch is constructed where the model trains off one example at a time and then uses different loss functions based on the task for that particular sample to compute the gradient and update the model accordingly.
\newline
\newline
\textbf{ScRNN:}
\newline
Similar to the MT-DNN \cite{liu2019multitask} , the ScRNN \cite{sakaguchi2016robsut} in first trained off the Penn Treebank dataset with synthetic noise added. 
\subsubsection{Finetuning}

\textbf{MT-DNN:}\newline
The MT-DNN model \cite{liu2019multitask} takes data which is pertinent to the use case for which it is being implemented in, and then finetunes the pretrained model to that specific scenario. These opportunities for finetuning as opposed to simply using the pretrained model allow for a greater flexibility and accuracy for the specific use case.

\section{Experiment Details}
\subsection{Adversarial Attacks}
After randomly assorting the sentence into a list of characters, a position is randomly chosen in order to determine where to perform the attack. With regards to a two character attack, the same procedure is repeated two get the adversarial example. In this experiment we only utilize the swap attack (where two adjacent characters' positions are switched), but the experiment also provides a framework to choose what attack to perform (delete, insert, shuffle, or trying all of them and finding which one compromises the accuracy the most)\footnote{\url{https://github.com/katchu11/Robustness-of-MT-DNNs/blob/master/attacks.py}}. As the attack position is chosen at random there is a possibility for the 2-char attack to be on the same word. However for sufficiently long sentences, this scenario is not frequently observed.

Samples of these attacks can be observed in \hyperref[tab:corrupted]{Table 2}, where the adversarially modified samples are shown in \textbf{bold}. This table also shows how 1-char and 2-char attacks are explored within the SNLI and SciTail datasets. 
\subsection{Architecture}
\subsubsection{MT-DNN}

We followed Liu et. al's  implementation of the MT-DNN \cite{liu2019multitask} . It used huggingface's PyTorch implementation of BERT for the parameter initialization \cite{wolf2019huggingfaces}. Then it used Adamax \cite{adamax} as the optimizer with a learning rate of 5e-5 and a batch size of
32 by following Devlin et al. \cite{BERT}. The maximum number of epochs was set to 5. A linear learning rate decay schedule with warm-up over
0.1 was used. We also set the dropout rate of all the task specific layers as 0.1, except 0.3 for MNLI and 0.05 for CoLA. To
avoid the exploding gradient problem, we clipped the gradient norm within 1. All the texts were tokenized using wordpieces, and were chopped to spans no longer than 512 tokens.

\subsubsection{ScRNN}
The input layer of ScRNN \cite{sakaguchi2016robsut} consists of a vector with length
of 76 (A-Z, a-z and 24 symbol characters). The hidden layer
units had size 650, and total vocabulary size was set to 10k.
We applied one type of noise to every word, but words with
numbers (e.g. 1980s) and short words (lengths less than 4) were
not subjected to jumbling, and therefore these words were
excluded in evaluation. We trained the model by running 5
epochs with (mini) batch size 20. We set the backpropagation through time (BPTT) parameter to 3 which means that the ScRNN updates
weights for previous two words $(x_{n-2}, x_{n-1})$ and the current word ($x_n$).

 \renewcommand{\arraystretch}{1.5}
\newpage

\begin{table}[h!]
\centering
\caption{Data from 1-character and 2-character attacks across various architectures}

\begin{tabular}{|c|c|c|c|c|c|c|}
\hline
\multicolumn{7}{|c|}{Format: 0 attacks, 1-char attack, 2-char attack}      \\ \hline
Architecture   & \multicolumn{3}{c|}{SNLI}  & \multicolumn{3}{c|}{SciTail} \\ \hline
BERT (Vanilla) & 90.50\% & 59.20\%  & 39.90\%  & 93.50\%  & 63.50\%   & 43.60\%  \\ \hline
MT-DNN         & 91.60\% & 54.96\% & 49.55\% & 95.00\%  & 78.83\%  & 62.76\% \\ \hline
MT-DNN + ScRNN & 91.50\% & 87.40\%  & 86.30\%  & 94.70\%  & 89.50\%   & 88.70\%  \\ \hline
\end{tabular}
\label{tab:data}
\end{table}
\renewcommand{\arraystretch}{1.5}
\begin{table}[h!]
\centering
\caption{Sample data showing corrupted samples}

\resizebox{\textwidth}{!}{%
\begin{tabular}{|c|c|c|}
\hline
Tasks                    & \# of attacks & Samples (Sentence 1 ||| Sentence 2)                                                                              \\ \hline
\multirow{2}{*}{SNLI}    & 1-char        & A soccer \textbf{gmae} with multiple males playing ||| A soccer game with multiple males playing.                         \\ \cline{2-3} 
                         & 2-char        & A soccer \textbf{gmae} with multiple males playing ||| Some men \textbf{rae} playing a sport.                                      \\ \hline
\multirow{2}{*}{SciTail} & 1-char        & The liver is \textbf{dviided} into the right lobe and left lobes ||| The gallbladder is near the right lobe of the liver. \\ \cline{2-3} 
                         & 2-char        & The liver is \textbf{dviided} into the right lobe and left lobes ||| The \textbf{gallbaldder} is near the right lobe of the liver. \\ \hline
\end{tabular}%
}
\label{tab:corrupted}
\end{table}

\begin{table}[h!]
\centering
\caption{Examples that passed the MT-DNN + ScRNN combo still failed at}

\resizebox{\textwidth}{!}{%
\begin{tabular}{|c|c|}
\hline
Datasets                 & Faultily identified sentences (Sentence 1 ||| Sentence 2)                                                                                                                                                                                          \\ \hline
\multirow{3}{*}{SNLI}    & \begin{tabular}[c]{@{}c@{}}Two women are embracing while holding to go \textbf{pcakages}. ||| \\ The sisters are hugging goodbye while holding to go packages after just eating lunch.\end{tabular}                                                         \\ \cline{2-2} 
                         & \begin{tabular}[c]{@{}c@{}}Man in a black suit, \textbf{wihte shitr} and black bowtie playing an instrument with the rest of his symphony surrounding him.||| \\ A person in a suit\end{tabular}                                                            \\ \cline{2-2} 
                         & \begin{tabular}[c]{@{}c@{}}A woman stands at a podium with a slide show behind her.||| \\ A woman is standing at a \textbf{pdoium}.\end{tabular}                                                                                                            \\ \hline
\multirow{3}{*}{SciTail} & \begin{tabular}[c]{@{}c@{}}\textbf{Gsaes} have neither definite volume nor shape ||| \\ Gas has no definite volume and no definite shape\end{tabular}                                                                                                       \\ \cline{2-2} 
                         & \begin{tabular}[c]{@{}c@{}}A lunar \textbf{elcipse} occurs when the Moon passes into the Earth's shadow ||| \\ When earth's shadow falls on the moon, the shadow causes a \textbf{lnuar} eclipse.\end{tabular}                                                       \\ \cline{2-2} 
                         & \begin{tabular}[c]{@{}c@{}}During \textbf{pohtosynthesis}, sunlight changes water and carbon dioxide into glucose and oxygen. ||| \\ Photosynthesis is described by this statement: carbon dioxide and water are turned into sugar and oxygen.\end{tabular} \\ \hline
\end{tabular}%
}
\label{tab:failures}
\end{table}
\subsection{Experimental Results}
As shown in \hyperref[tab:data]{Table 1}, we can see that the as we perform character level attacks on the defenseless architectures (BERT, and the MT-DNN), the accuracy significantly reduces with a drop off of 31.3\% and 30.0\% across datasets after one attack for BERT, and a drop off of 36.64\% and 16.67\% across datasets for the MT-DNN. When using a two character attack, the accuracy drops off by 50.6\% and 49.9\% across datasets for the BERT model, and by 42.05\% and 32.24\% across datasets for the MT-DNN model. This data shows the already pre-existing resilience to attacks from the MT-DNN model due to it's cross training, as both drop offs in accuracy from the MT-DNN are still lower than drop offs in accuracy from BERT.
\newline
\newline
However when prefixing a Semicharacter RNN as a pre-processing step for the data, we observe a significant "restoration" in accuracy. When utilizing an ScRNN along with an MT-DNN, the accuracy is restored by 32.44\% and 10.67\% across datasets for 1 character attacks, and by 36.75\% and by 25.94\% for 2 character attacks across datasets.
\newline
\newline
\subsubsection{Error Analysis}
When observing corrupted samples (shown in  \hyperref[tab:corrupted]{Table 2} and  \hyperref[tab:failures]{Table 3}) however that still passed through the MT-DNN + ScRNN combination, it can be seen that they are mainly samples with \emph{domain-specific} words which throw off the prediction. Most of the errors occurred when domain-specific words such as "Gases" or "Photosynthesis" were misspelled (incorrect samples shown in \textbf{bold} for both tables). This shows that the models performance could potentially be increased through the use of a domain-specific training step for the ScRNN. This could be included a priori to deployment by either training the ScRNN in parallel to the MT-DNN, or separately (before or after). 

These results demonstrate that the MT-DNN is inherently more resilient to attacks than BERT due to it's multi-task learning, but that it's accuracy can also be restored significantly when utilizing a domain adaptable defense.


\section{Conclusion}
Through our training we effectively show that the MT-DNN is a very versatile architecture for NLU tasks. Although it can be severely compromised to basic adversarial attacks, through the use of a domain adaptable defense, it's accuracy can be almost completely restored in order to ensure optimal performance. There is however room for improvement to perhaps integrate more than one system of adversarial resistance. This model can also be improved by adding a domain-specific fine-tuning step to the ScRNN to ensure that words that aren't especially common for a certain domain are still noted.
\newpage
\bibliographystyle{unsrt}  


\end{document}